\pdfoutput=1
\documentclass[10pt,twocolumn,letterpaper]{article}
\usepackage{iccv}
\usepackage{times}
\usepackage{epsfig}
\usepackage{graphicx}
\usepackage{amsmath}
\usepackage{amssymb}
\usepackage{hyperref}

\iccvfinalcopy 

\def\iccvPaperID{4488} 

\ificcvfinal\fi

\begin{document}

\title{Compositional 3D Scene Generation using Locally Conditioned Diffusion}

\author{Ryan Po\\
Stanford University\\
{\tt\small rlpo@stanford.edu}
\and
Gordon Wetzstein\\
Stanford University\\
{\tt\small gordon.wetzstein@stanford.edu}
}

\maketitle
\ificcvfinal\thispagestyle{empty}\fi

\begin{abstract}
Designing complex 3D scenes has been a tedious, manual process requiring domain expertise. Emerging text-to-3D generative models show great promise for making this task more intuitive, but existing approaches are limited to object-level generation. We introduce \textbf{locally conditioned diffusion} as an approach to compositional scene diffusion, providing control over semantic parts using text prompts and bounding boxes while ensuring seamless transitions between these parts. We demonstrate a score distillation sampling--based text-to-3D synthesis pipeline that enables compositional 3D scene generation at a higher fidelity than relevant baselines.
\end{abstract}

\newcommand\blfootnote[1]{%
  \begingroup
  \renewcommand\thefootnote{}\footnote{#1}%
  \addtocounter{footnote}{-1}%
  \endgroup
}

\blfootnote{Project page: \color{magenta}\url{http://www.ryanpo.com/comp3d}}

\section{Introduction}
\begin{figure*}[t!]
    \vspace{-1em}
    \centering
    \includegraphics[width=\linewidth]{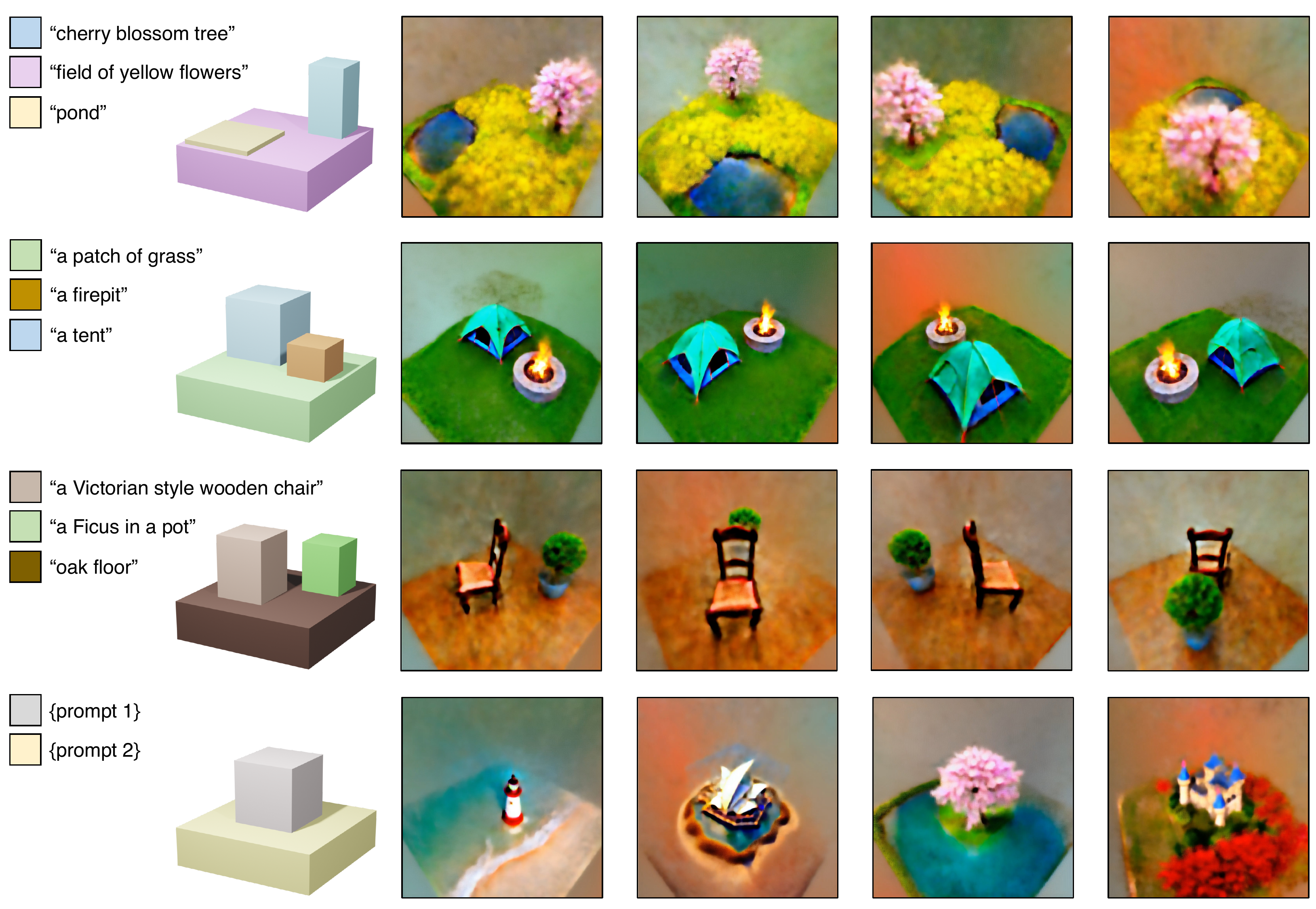}
    \vspace{-1em}
    \caption{\textbf{Results of our method.} Given user-input bounding boxes with corresponding text prompts, our method is able to generate high-quality 3D scenes that adhere to the desired layout with seamless transitions. Our locally conditioned diffusion method blends multiple objects into a single coherent scene, while simultaneously providing control over the size and position of individual scene components. Text prompts for bottom row (from left): (1) ``a lighthouse'' and (2) ``a beach''; (1) ``the Sydney Opera House'' and (2) ``a desert''; (1) ``a cherry blossom tree'' and (2) ``a lake''; (1) ``a small castle'' and (2) ``a field of red flowers''. Videos of our results can be found in the supplementary materials.}
    \label{fig:main_results}
    \vspace{-1em}
\end{figure*}

Traditionally, 3D scene modelling has been a time-consuming process exclusive to those with domain expertise. While a large bank of 3D assets exists in the public domain, it is quite rare to find a 3D scene that fits the user’s exact specifications. For this reason, 3D designers often spend hours to days modelling individual 3D assets and composing them together into a scene. To bridge the gap between expert 3D designers and the average person, 3D generation should be made simple and intuitive while maintaining control over its elements (e.g., size and position of individual objects).

Recent work on 3D generative models has made progress towards making 3D scene modelling more accessible. 3D-aware generative adversarial networks (GANs) ~\cite{Liao2020unsupervised-3d-synthesis,Wu20163d-latent-gan,Nguyen-Phuoc2019hologan,Gadelha20173d-shape-induction,Niemeyer2021giraffe,Liu2020neural,chan2020pi, graf, Meng2021gnerf, Kosiorek2021nerf-vae, Rebain2022lolnerf,Chan2022eg3d,gu2021stylenerf,Zhou2021CIPS3D,Or-El2022style-sdf,zheng2022sdfstylegan} have shown promising results for 3D object synthesis, demonstrating elementary progress towards composing generated objects into scene~\cite{Niemeyer2020GIRAFFE,Xu2022DisCoSceneSD}. However, GANs are specific to an object category, limiting the diversity of results and making scene-level text-to-3D generation challenging. In contrast, text-to-3D generation~\cite{Poole2022DreamFusionTU, Lin2022Magic3DHT, Wang2022ScoreJC} using diffusion models can generate 3D assets from a wide variety of categories via text prompts~\cite{Balaji2022eDiffITD, Rombach2021HighResolutionIS, Saharia2022PhotorealisticTD}. Existing work leverages strong 2D image diffusion priors trained on internet-scale data, using a single text prompt to apply a global conditioning on rendered views of a differentiable scene representation. Such methods can generate high-quality object-centric generations but struggle to generate scenes with multiple distinct elements. Global conditioning also limits controllability, as user input is constrained to a single text prompt, providing no control over the layout of the generated scene.

We introduce locally conditioned diffusion, a method for compositional text-to-image generation using diffusion models. Taking an input segmentation mask with corresponding text prompts, our method selectively applies conditional diffusion steps to specified regions of the image, generating outputs that adhere to the user specified composition. We also achieve compositional text-to-3D scene generation by applying our method to a score distillation sampling--based text-to-3D generation pipeline. Our proposed method takes 3D bounding boxes and text prompts as input and generates coherent 3D scenes while providing control over size and positioning of individual assets. Specifically, our contributions are the following:
\begin{itemize}
    \item We introduce \textbf{locally conditioned diffusion}, a method that allows greater compositional control over existing 2D diffusion models.
    \item We introduce a method for compositional 3D synthesis by applying locally conditioned diffusion to a score distillation sampling--based 3D generative pipeline.
    \item We introduce key camera pose sampling strategies, crucial for compositional 3D generation.
\end{itemize}

\section{Related work}

\paragraph{2D diffusion models.} Advances in large-scale 2D diffusion models trained on internet-scale data \cite{Dhariwal2021DiffusionMB, Nichol2021GLIDETP, Rombach2021HighResolutionIS, Radford2021LearningTV, Ramesh2022HierarchicalTI, Saharia2022PhotorealisticTD, Schuhmann2022LAION5BAO} have allowed generation of high-quality images that stay accurate to complex text prompts. While text-conditioned diffusion models excel at reproducing the semantics of a prompt, compositional information is usually ignored. Variants of existing methods \cite{Rombach2021HighResolutionIS} instead condition their models with semantic bounding boxes. This change allows greater control over the composition of the generated image. However, bounding-box-conditioned models must be trained with annotated image data \cite{Caesar2016COCOStuffTA}. These datasets are often much more limited in size, which restricts the diversity of the resulting diffusion model. Our locally conditioned diffusion approach leverages pre-trained text-conditioned 2D diffusion models to generate high-quality images with better compositional control without restricting the complexity of user-provided text-prompts.
\begin{figure*}[t!]
    \vspace{-1em}
    \centering
    \includegraphics[width=\linewidth]{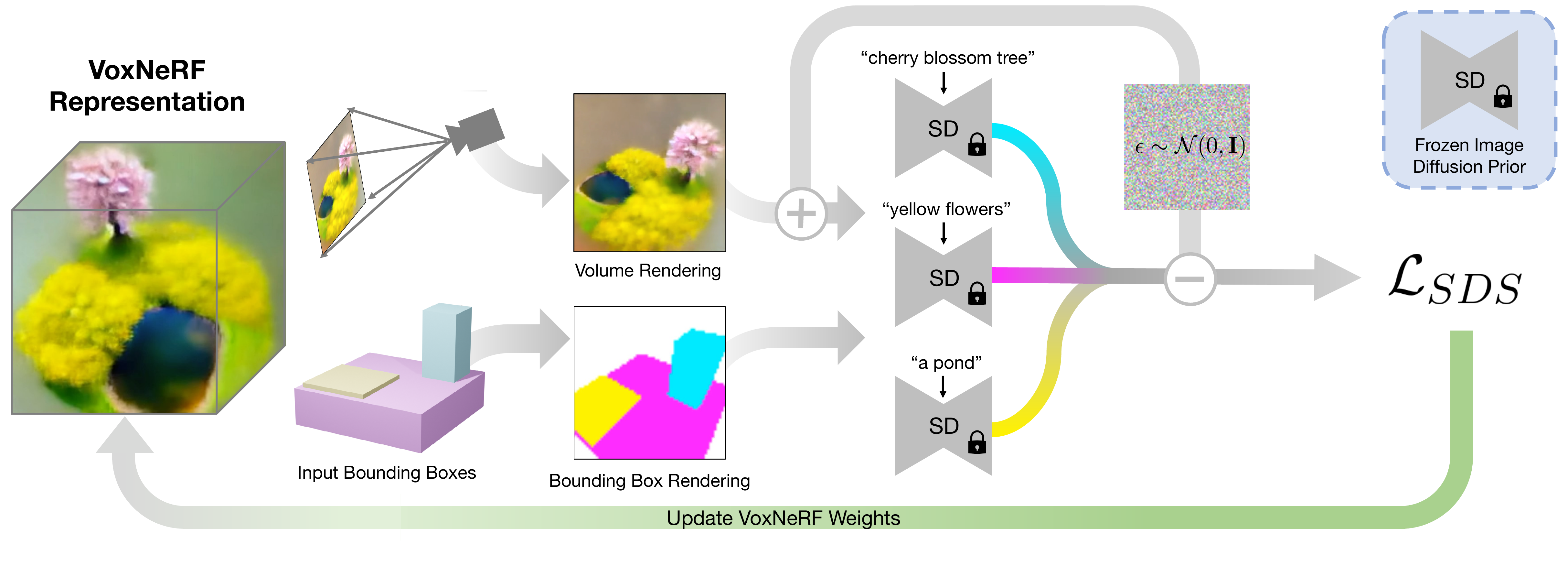}
    \vspace{-1em}
    \caption{\textbf{Overview of our method.} We generate text-to-3D content using a score distillation sampling--based pipeline. A latent diffusion prior is used to optimize a Voxel NeRF representation of the 3D scene. The latent diffusion prior is conditioned on a bounding box rendering of the scene, where a noise estimation on the image is formed for every input text prompt, and denoising steps are applied based on the segmentation mask provided by the bounding box rendering.}
    \label{fig:3d_pipeline}
\end{figure*}
\paragraph{Compositional image generation.} Recent work found that Energy-Based Models (EBMs) \cite{Du2020CompositionalVG, Du2019ImplicitGA, LeCun2006ATO, Gao2020LearningEM, Grathwohl2020LearningTS} tend to struggle with composing multiple concepts into a single image \cite{Du2020CompositionalVG, Liu2022CompositionalVG}. Noting that EBMs and diffusion models are functionally similar, recent work improves the expressivity of diffusion by borrowing theory from EBMs. For example, \cite{Liu2022CompositionalVG} achieves this by composing gradients from denoisers conditioned on separate text-prompts in a manner similar to classifier-free guidance as proposed by \cite{Dhariwal2021DiffusionMB}. Existing work, such as Composable-Diffusion \cite{Liu2022CompositionalVG}, however, apply composition to the entire image, offering no control over the position and size of different concepts. Our locally conditioned diffusion approach selectively applies denoising steps over user-defined regions, providing increased compositional control for image synthesis while ensuring seamless transitions.

\paragraph{Text-to-3D diffusion models.} Recent advances in 2D diffusion models have motivated a class of methods for performing text-to-3D synthesis. Existing methods leverage 2D diffusion models trained on internet-scale data to achieve text-to-3D synthesis. Notably, DreamFusion \cite{Poole2022DreamFusionTU} with Imagen \cite{Saharia2022PhotorealisticTD}, Score Jacobian Chaining (SJC) \cite{Wang2022ScoreJC} with StableDiffusion \cite{Rombach2021HighResolutionIS} and Magic3D \cite{Lin2022Magic3DHT} with eDiff-I \cite{Balaji2022eDiffITD} and StableDiffusion \cite{Rombach2021HighResolutionIS}. Previous methods \cite{Poole2022DreamFusionTU, Wang2022ScoreJC, Lin2022Magic3DHT, Metzer2022LatentNeRFFS} perform 3D synthesis by denoising rendered views of a differentiable 3D representation. This process is coined Score Distillation Sampling (SDS) by the authors of DreamFusion \cite{Poole2022DreamFusionTU}. Intuitively, SDS ensures that all rendered views of the 3D representation resemble an image generated by the text-conditioned 2D diffusion model. Current methods are able to generate high quality 3D assets from complex text prompts. However, they are unable to create 3D scenes with specific compositions. Our proposed method enables explicit control over size and position of scene components.
\section{Diffusion preliminaries}
Recent work has shown that diffusion models can achieve state-of-the-art quality for image generation tasks~\cite{Dhariwal2021DiffusionMB}. Specifically, Denoising Diffusion Probabilistic Models (DDPMs) implement image synthesis as a denoising process. DDPMs begin from sampled Gaussian noise $x_T$ and apply $T$ denoising steps to create a final image $x_0$. 
The forward diffusion process $q$ is modelled as a Markov chain that gradually adds Gaussian noise to a ground truth image according to a predetermined variance schedule $\beta_1, \beta_2, \dots, \beta_T$
\begin{equation}
    q(x_t | x_{t-1}) = \mathcal{N} \left(x_t ; \sqrt{1-\beta_t}x_{t-1} , \beta_t \mathbf{I} \right)
\end{equation}
The goal of DDPMs is to train a diffusion model to revert the forward process. Specifically, a function approximator $\boldsymbol{\epsilon}_\phi$ is trained to predict the noise $\boldsymbol{\epsilon}$ contained in a noisy image $x_t$ at step $t$. $\boldsymbol{\epsilon}_\phi$ is typically represented as a convolutional neural network characterised by its parameters $\phi$. Most successful models \cite{Dhariwal2021DiffusionMB, Ho2020DenoisingDP, Saharia2021ImageSV} train their models using a simplified variant of the variational lower bound on the data distribution:
\begin{equation}
    \mathcal{L}_\textrm{DDPM} \! = \! \mathbb{E}_{t,x, \boldsymbol{\epsilon}} \left[ \left\| \boldsymbol{\epsilon} - \boldsymbol{\epsilon}_\phi \left(x_t, t \right) \right\|^2 \right]\!\!
    \label{eq:loss}
\end{equation}
with $t$ uniformly sampled from $\{1,\dots,T\}$.
The resulting update step for obtaining a sample for $x_{t-1}$ from $x_t$ is then
\begin{equation}
    x_{t-1} = x_t - \frac{1-\alpha_t}{\sqrt{1-\bar{\alpha}_t}} \epsilon_\phi(x_t,t) + \frac{1-\bar{\alpha}_{t-1}}{1-\bar{\alpha}_t}\beta_t \mathcal{N}(0,\mathbf{I})
\end{equation}
where $\bar{\alpha}_t = \prod_{s=1}^t \alpha_s$, $\alpha_t = 1-\beta_t$. 

Text-to-image diffusion models build upon the above theory to introduce conditional diffusion processes using classifier-free guidance \cite{Ho2022ClassifierFreeDG}. Given a condition $y$, usually represented as a text prompt, a diffusion model $\boldsymbol{\epsilon}_\phi(x_t, t, y)$ is trained to predict noise in an image as shown in Eq. \ref{eq:loss}. During training, conditioning $y$ is randomly dropped out, leaving the diffusion model to predict noise without it. At inference, noise prediction is instead represented by:
\begin{equation}
    \hat{\boldsymbol{\epsilon}}_\phi(x_t, t, y) = \boldsymbol{\epsilon}_\phi(x_t, t, \emptyset) + s\Bigl(\boldsymbol{\epsilon}_\phi(x_t, t, y) - \boldsymbol{\epsilon}_\phi(x_t, t, \emptyset)\Bigr)
\end{equation}
Where $s$ is a user-defined constant controlling the degree of guidance and $\boldsymbol{\epsilon}(x_t, t, \emptyset)$ represents the noise prediction without conditioning.

\begin{figure*}[t]
    \centering
    \includegraphics[width=\linewidth]{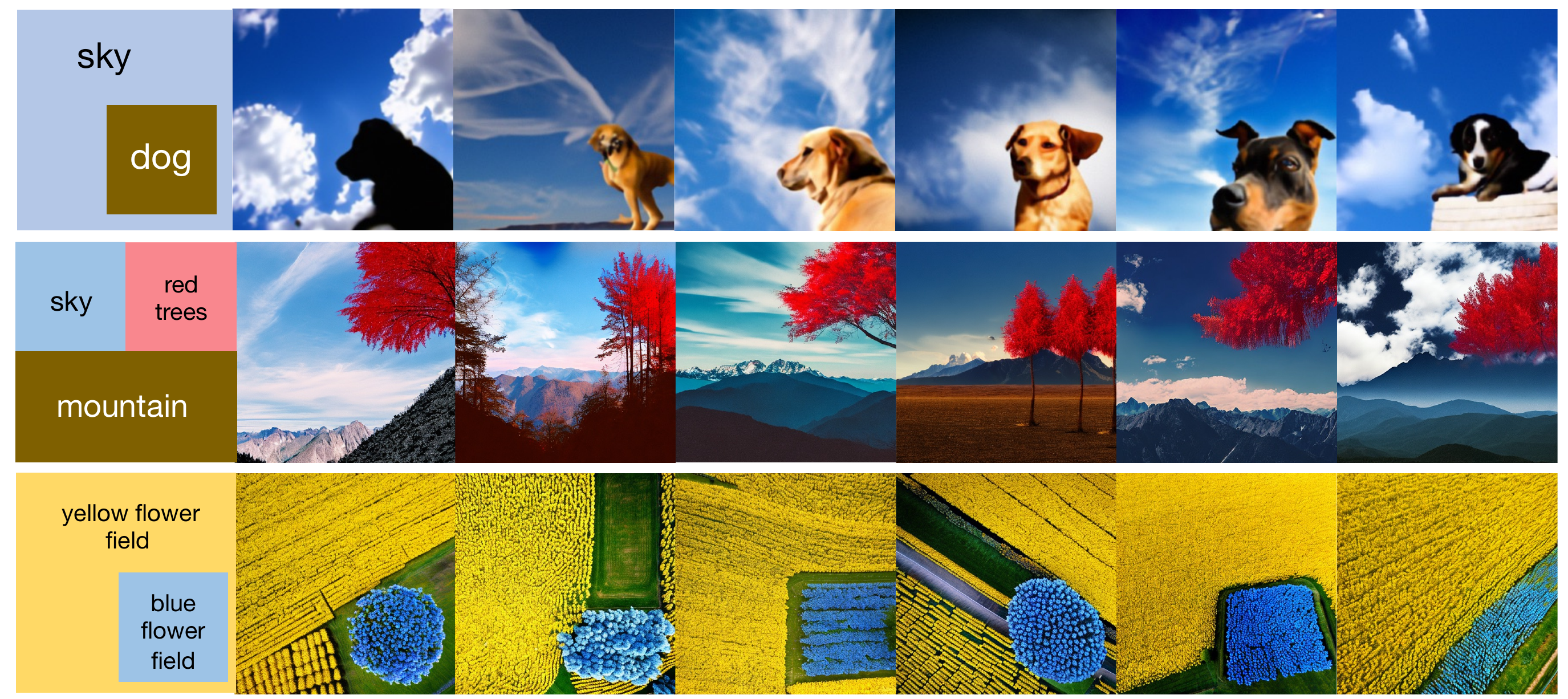}
    \vspace{-1em}
    \caption{\textbf{2D locally conditioned diffusion results.} Given coarse segmentation masks as input, our method is able to generate images that follow the specified layout while ensuring seamless transitions. Results in the first row are generated using GLIDE \cite{Nichol2021GLIDETP}, while the second and third rows show results generated using StableDiffusion \cite{Rombach2021HighResolutionIS}.}
    \label{fig:2d_results}
\end{figure*}

\begin{figure*}[t!]
    \centering
    \includegraphics[width=\linewidth]{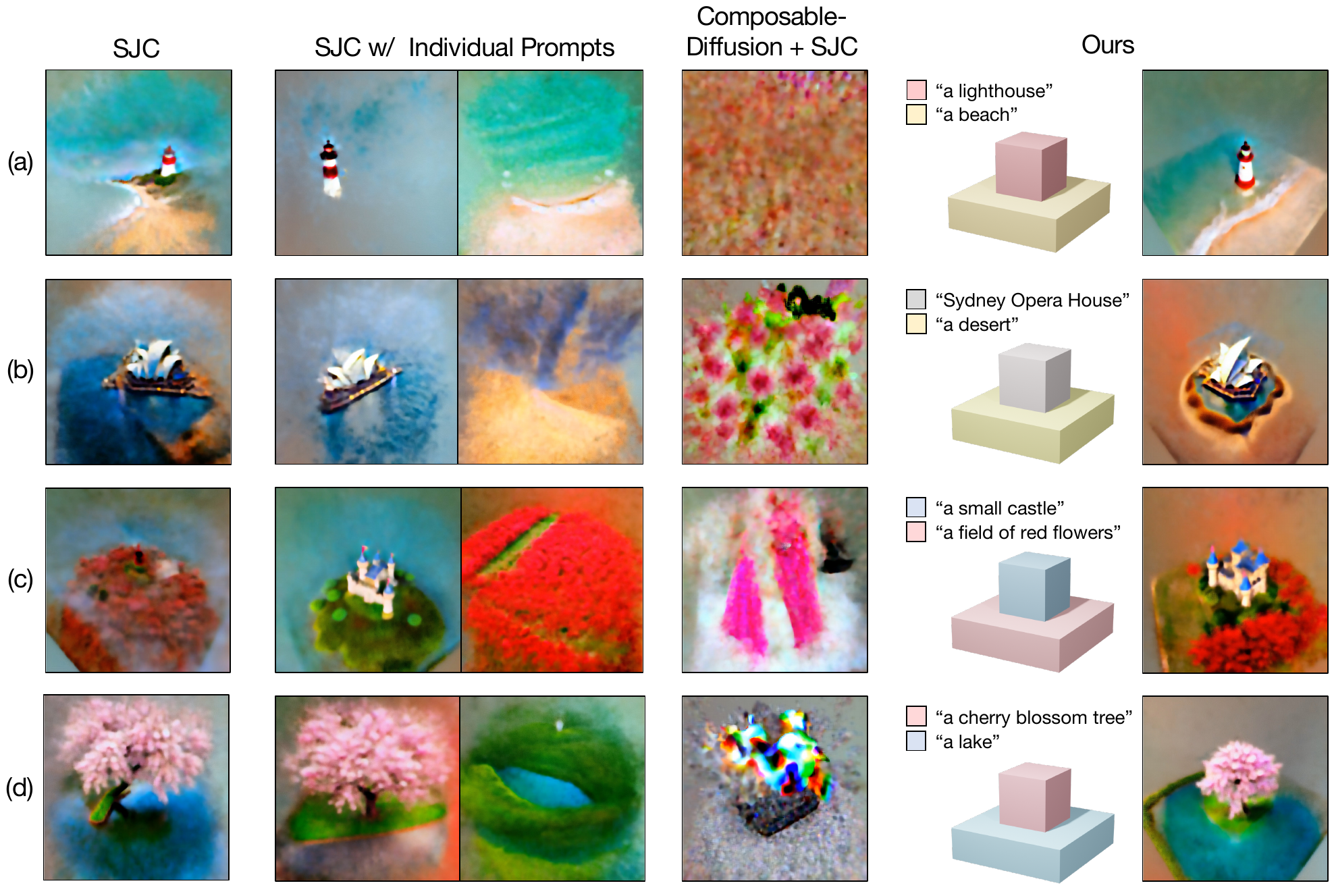}
    \vspace{-1em}
    \caption{\textbf{Baseline comparisons.} Left to right: (i) SJC results using a single text prompt, (ii) SJC generating each scene component independently, (iii) SJC combined with Composable-Diffusion \cite{Liu2022CompositionalVG}, and (iv) our method with corresponding bounding boxes and text prompts. Generations for each row use the text prompts listed on the right. Results in the first column are generated by combining individual text prompts with the connecting phrase ``in the middle of'', e.g. ``a lighthouse in the middle of a beach''. Our method successfully composes different objects into a coherent scene while following the user input bounding boxes.}
    \label{fig:baselines}
    \vspace{-1em}
\end{figure*}

\section{Locally conditioned diffusion}
We introduce \textbf{locally conditioned diffusion} as a method for providing better control over the composition of images generated by text-conditioned diffusion models. The key insight of our method is that we can selectively apply denoising steps conditioned on different text prompts to specific regions of an image.

Given a set of text prompts $\{y_1, \dots, y_P\}$, classifier-free guidance \cite{Ho2022ClassifierFreeDG} provides a method for predicting denoising steps conditioned on $y_i$:
\begin{equation}
    \hat{\boldsymbol{\epsilon}}_\phi(x_t, t, y_i) = \boldsymbol{\epsilon}_\phi(x_t, t, \emptyset) + s\Bigl(\boldsymbol{\epsilon}_\phi(x_t, t, y_i) - \boldsymbol{\epsilon}_\phi(x_t, t, \emptyset)\Bigr)
\end{equation}
Using a user-defined semantic segmentation mask $m$, where each pixel $m[j]$ has integer value $[1,P]$, the overall noise prediction can then be represented by selectively applying noise predictions to each labelled image patch:
\begin{equation}
    \hat{\boldsymbol{\epsilon}}_\phi(x_t, t, y_{1:P}, m) = \sum_{i=1}^P 
    \mathds{1}_{i}(m) \odot\hat{\boldsymbol{\epsilon}}_\phi(x_t, t, y_i)
    \label{sem_eps}
\end{equation}
Where $\mathds{1}_{i}(m)$ is the indicator image with equivalent dimensionality as $m$ and
\begin{equation} 
\mathds{1}_{i}(m)[j] = 
    \begin{cases}
    1,& \text{if } m[j] = i\\
    0,              & \text{otherwise}
\end{cases}
\end{equation}
The proposed locally conditioned diffusion method is summarized in Algorithm \ref{alg:semdiff}. 

\begin{algorithm}
\caption{Locally conditioned diffusion}\label{alg:semdiff}
\begin{algorithmic}
\Require Diffusion models $\hat{\boldsymbol{\epsilon}}_\phi(x_t, t, y_i)$, guidance scale $s$, semantic mask $m$
\State $x_T \sim \mathcal{N}(0, I)$ \Comment{Initialize Gaussian noise image} 
\For{$t = T,\dots,1$}
    \State $\epsilon_i \gets \hat{\boldsymbol{\epsilon}}_\phi(x_t, t, y_i)$ \Comment{Individual noise predictions} 
    \State $\epsilon \gets \hat{\boldsymbol{\epsilon}}_\phi(x_t, t, \emptyset)$ \Comment{Unconditional noise prediction} 
    \State $\epsilon_{\textrm{sem}} \gets \sum_{i=1}^P \mathds{1}_i(m) \odot s(\epsilon_i - \epsilon)$ \Comment{Combine noise predictions} 
    \State $x_{t-1} = \textrm{Update}(x_t, \epsilon_{\textrm{sem}})$ \Comment{Apply denoising step}
\EndFor
\end{algorithmic}
\end{algorithm}Although a large proportion of noise predictions are not used, in practice only one diffusion model $\boldsymbol{\epsilon}_\phi$ is queried. All calls to the model for each unique text-conditioning $y_i$ can be batched together for increased efficiency. 

Our locally conditioned diffusion method generates high-fidelity 2D images that adhere to the given semantic segmentation masks. Note that, while each segment of the image is locally conditioned, there are no visible seams in the resulting image and transitions between differently labelled regions are smooth, as shown in Fig. \ref{fig:2d_results} (see Sec. \ref{experiments} for more details).

\section{Compositional 3D synthesis}

To make compositional text-to-3D synthesis as simple as possible, our method takes 3D bounding boxes with corresponding text prompts as input. The goal of our method is to generate 3D scenes that contain objects specified by the text prompts while adhering to the specific composition provided by the input bounding boxes. In this section, we describe our method and how we apply locally conditioned diffusion in 2D to enable controllable generation in 3D.
\paragraph{Text-to-3D with Score Distillation Sampling.} Our method builds off existing SDS-based text-to-3D methods \cite{Lin2022Magic3DHT, Poole2022DreamFusionTU, Wang2022ScoreJC}. SDS-based methods leverage a 3D scene representation parameterized by $\theta$ is differentiably rendered at a sampled camera pose, generating a noised image $g(\theta)$ which is passed into an image diffusion prior. Our method builds off SJC \cite{Wang2022ScoreJC}, therefore we follow their pipeline, using a Voxel NeRF \cite{Chen2022TensoRFTR, Liu2020NeuralSV, Sitzmann2018DeepVoxelsLP, Sun2021DirectVG, Yu2021PlenoxelsRF} representation and a volumetric renderer. The image diffusion prior provides the gradient direction to update scene parameters $\theta$. 
\begin{equation}
    \nabla_\theta \mathcal{L}_\textrm{SDS} (\phi,g(\theta)) = \mathbb{E}_{t,\epsilon} \Bigl[ w(t)(\hat{\epsilon}(x_t, y, t) - \epsilon) \frac{\partial x}{\partial \theta}\Bigr]
\end{equation}
This process is repeated for randomly sampled camera poses, as the text-conditioned image diffusion prior pushes each rendered image towards high density regions in the data distribution. Intuitively, SDS ensures images rendered from all camera poses resembles an image generated by the text-conditioned diffusion prior.

\paragraph{Bounding-box-guided text-to-3D synthesis.} To achieve text-to-3D scene generations that adhere to user input bounding boxes, our method takes the standard SDS-based pipeline and conditions the image diffusion prior with renderings of the input bounding boxes. Specifically, our method works as follows. First, a random camera pose is sampled and a volume rendering of the 3D scene model is generated, we call this image $x_t$. Using the same camera pose, a rendering of the bounding boxes is also generated, we call this image $m$. This image is a segmentation mask, where each pixel contains an integer value corresponding to a user input text prompt. The volume rendering is then passed in to the image diffusion prior which provides the necessary gradients for optimizing the 3D scene representation. However, instead of conditioning the image diffusion prior on a single text prompt, we generate denoising steps for all text prompts with corresponding bounding boxes visible from the sampled camera pose. We then selectively apply these denoising steps to the image based on the segmentation mask $m$, and backpropagate the gradients to the 3D scene as usual. This is equivalent to applying the noise estimator described in Eq. \ref{sem_eps} to the SDS gradient updates. 
\begin{equation}
    \nabla_\theta \mathcal{L}_\textrm{SDS} (\phi,g(\theta),m) = \mathbb{E}_{t,\epsilon} \Bigl[ w(t)(\hat{\boldsymbol{\epsilon}}_\phi(x_t, t, y_{1:P}, m) - \epsilon) \frac{\partial x}{\partial \theta}\Bigr]
\end{equation}
While previous SDS-based text-to-3D methods ensure all rendered views of the 3D scene lie in the high probability density regions in the image prior conditioned on a single text prompt, our method ensures that all rendered views also align with the rendered bounding box segmentation masks. An overview of our method is provided in Fig. \ref{fig:3d_pipeline}.
\paragraph{Object-centric camera pose sampling.} As discussed in prior work \cite{Poole2022DreamFusionTU, Lin2022Magic3DHT, Wang2022ScoreJC}, high classifier-free guidance weights are crucial for SDS methods to work. While image generation methods typically use guidance weights in the range of [2, 50], methods such as DreamFusion use guidance weights up to 100 \cite{Poole2022DreamFusionTU}. Using a high guidance scale leads to mode-seeking properties which is desirable in the context of SDS-based generation. However, mode-seeking properties in image diffusion priors have the tendency of generating images with the object at the center of the image. When applying high guidance weights to locally conditioned diffusion, it is possible for the resulting image to ignore semantic regions that are off center, since mode-seeking behaviour of the diffusion model expects the object described by the text prompt to be at the center of the image, while the semantic mask only applies gradients from off-centered regions. In the context of our method, this mode-seeking behavior causes off-centered bounding box regions to become empty.

We combat this effect using \textit{object-centric camera pose sampling}. While existing works \cite{Poole2022DreamFusionTU, Lin2022Magic3DHT, Wang2022ScoreJC} sample camera poses that are always pointed at the origin of the 3D scene model, in our method, we randomly sample camera poses that point at the center of each object bounding box instead. This means that during optimization of the 3D scene, each bounding box region will have the chance at appearing at the center of the image diffusion prior.

\paragraph{Locally conditioned diffusion with latent diffusion models.} Existing SDS-based methods, such as DreamFusion \cite{Poole2022DreamFusionTU} and Magic3D \cite{Lin2022Magic3DHT}, leverage image diffusion priors in their method \footnote{In Magic3D, a latent diffusion prior is also used, but the gradient of the encoder in the latent diffusion model is provided to convert gradient updates in the latent space back to the image space.}. While SJC \cite{Wang2022ScoreJC} uses a very similar methodology, their method actual employs a latent diffusion prior in the form of StableDiffusion \cite{Rombach2021HighResolutionIS}. Therefore, volume renderings of the 3D scene lies in the latent space instead of the image space. Note that previous work \cite{Park2022ShapeGuidedDW} has shown that the latent space is essentially a downsampled version of the image space, meaning we are still able to apply locally conditioned diffusion to the latent space.
\begin{figure*}[t]
    \centering
    \includegraphics[width=\linewidth]{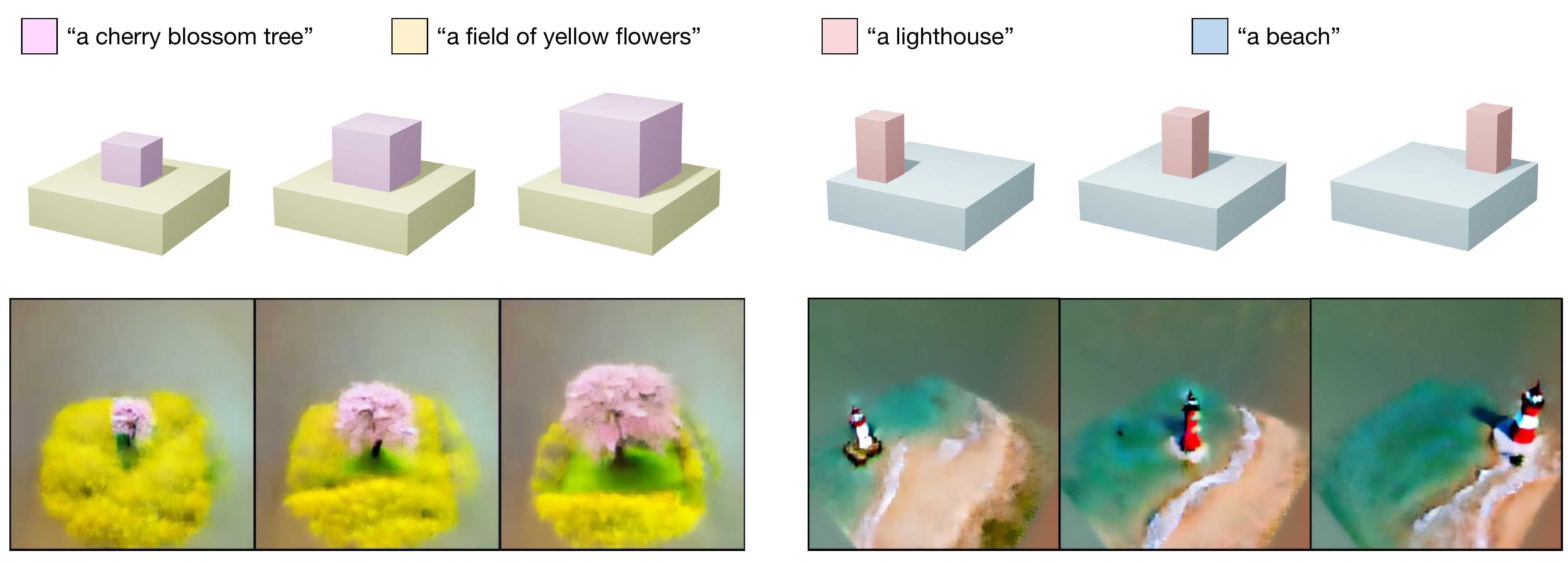}
    \vspace{-1em}
    \caption{\textbf{Size and position control.} Our method provides size and position control of individual scene components through user-defined bounding boxes. Our method provides fine-grained control over scene composition while ensuring each components blends seamlessly into the overall scene.}
    \label{fig:size_and_position}
    \vspace{-1em}
\end{figure*}
\section{Experiments}\label{experiments}
We show qualitative results on compositional text-to-2D and text-to-3D generation. For 3D results, we mainly compare against SJC \cite{Wang2022ScoreJC} as it is the best-performing publicly-available text-to-3D method. We also implemented a version of SJC that leverages Composable-Diffusion \cite{Liu2022CompositionalVG} as an additional baseline.

\subsection{Compositional 2D results}

\paragraph{Implementation details.} We apply our locally conditioned diffusion method to existing text-conditioned diffusion models: GLIDE~\cite{Nichol2021GLIDETP} and StableDiffusion~\cite{Rombach2021HighResolutionIS}. We use pre-trained models provided by the authors of each respective paper to implement locally conditioned diffusion. Each image sample takes 10--15 seconds to generate on an NVIDIA A100 GPU, where duration varies according to the number of distinct semantic regions provided. Note that sampling time increases sub-linearly with respect to number of regions/prompts, this is because calls to the same model for each text-conditioning can be done in a single batch.

\paragraph{Qualitative results.} We provide qualitative examples in  Fig. \ref{fig:2d_results}. Our method is able to generate high-fidelity images that adhere to the input semantic masks and text prompts. Note that our method does not aim at generating images that follow the exact boundaries of the input semantic masks, instead it strives to achieve seamless transitions between different semantic regions. A key advantage of locally conditioned diffusion is that it is agnostic to the network architecture. We demonstrate this by showing that our method works on two popular text-to-image diffusion models GLIDE \cite{Nichol2021GLIDETP} and StableDiffusion \cite{Rombach2021HighResolutionIS}.

\subsection{Compositional 3D results}

\paragraph{Implementation details.} Our compositional text-to-3D method builds upon the SJC \cite{Wang2022ScoreJC} codebase. Following SJC, we use a Voxel NeRF to represent the 3D scene model and StableDiffusion \cite{Rombach2021HighResolutionIS} as the diffusion prior for SDS-based generation. The Voxel NeRF representing the 3D scene is set to a resolution of $100^3$. This configuration uses $\approx10$ GB of GPU memory. The original SJC method uses an emptiness loss scheduler to improve the quality of generated scenes. Our method also leverages this emptiness loss; please refer to the original SJC \cite{Wang2022ScoreJC} for more details.

\paragraph{Qualitative results.} We provide qualitative examples of compositional text-to-3D generations with bounding box guidance in Fig. \ref{fig:main_results}. Notice that our method is able to generate coherent 3D scenes using simple bounding boxes with corresponding text prompts. Our method generates results that adhere to the input bounding boxes, allowing users to edit the size and position of individual scene components before generation. Fig. \ref{fig:size_and_position} shows generated results of the same scene prompts with differing bounding box sizes and positions. Note that our method is able to adapt to the user's input and generate scenes with varying compositions.

\paragraph{Baseline comparisons.} We compare our method to different variants of SJC \cite{Wang2022ScoreJC}. Namely, (i) SJC generations using a single prompt for the entire scene, (ii) individual SJC generations for each scene component, and (iii) an implementation of Composable-Diffusion \cite{Liu2022CompositionalVG} combined with SJC. Although DreamFusion \cite{Poole2022DreamFusionTU} and Magic3D \cite{Lin2022Magic3DHT} have also shown to generate high-quality results, both models leverage image diffusion priors (Imagen \cite{Saharia2022PhotorealisticTD} and eDiff-I \cite{Balaji2022eDiffITD}) that are not publicly available. However, it is important to note that our method can theoretically be applied to any SDS-based method. This can be achieved by replacing the image diffusion model in DreamFusion \cite{Poole2022DreamFusionTU} and Magic3D \cite{Lin2022Magic3DHT} with the locally conditioned method described above.

We provide qualitative results for our method and each baseline in Fig. \ref{fig:baselines}. In our experiments we attempt to compose two scene components into a coherent scene. Specifically, we choose an object-centric prompt that describes individual objects, paired with a scene-centric prompt that describes a background or an environment. 

We observe that SJC fails to capture certain scene components when composing multiple scene components into a single prompt. Our method is able to capture individual scene components while blending them seamlessly into a coherent scene. 

For object-centric prompts, SJC is able to create high-quality 3D generations. However, scene-centric prompts such as ``a desert'' or ``a beach'' end up generating dense volumes that resemble the text-prompt when rendered from different angles, but fail to reconstruct reasonable geometry. By defining bounding boxes for each scene component, our method provides coarse guidance for the geometry of the scene, this helps generate results with fewer ``floater'' artifacts.
\begin{figure}[tb!]
    \centering
    \includegraphics[width=\linewidth]{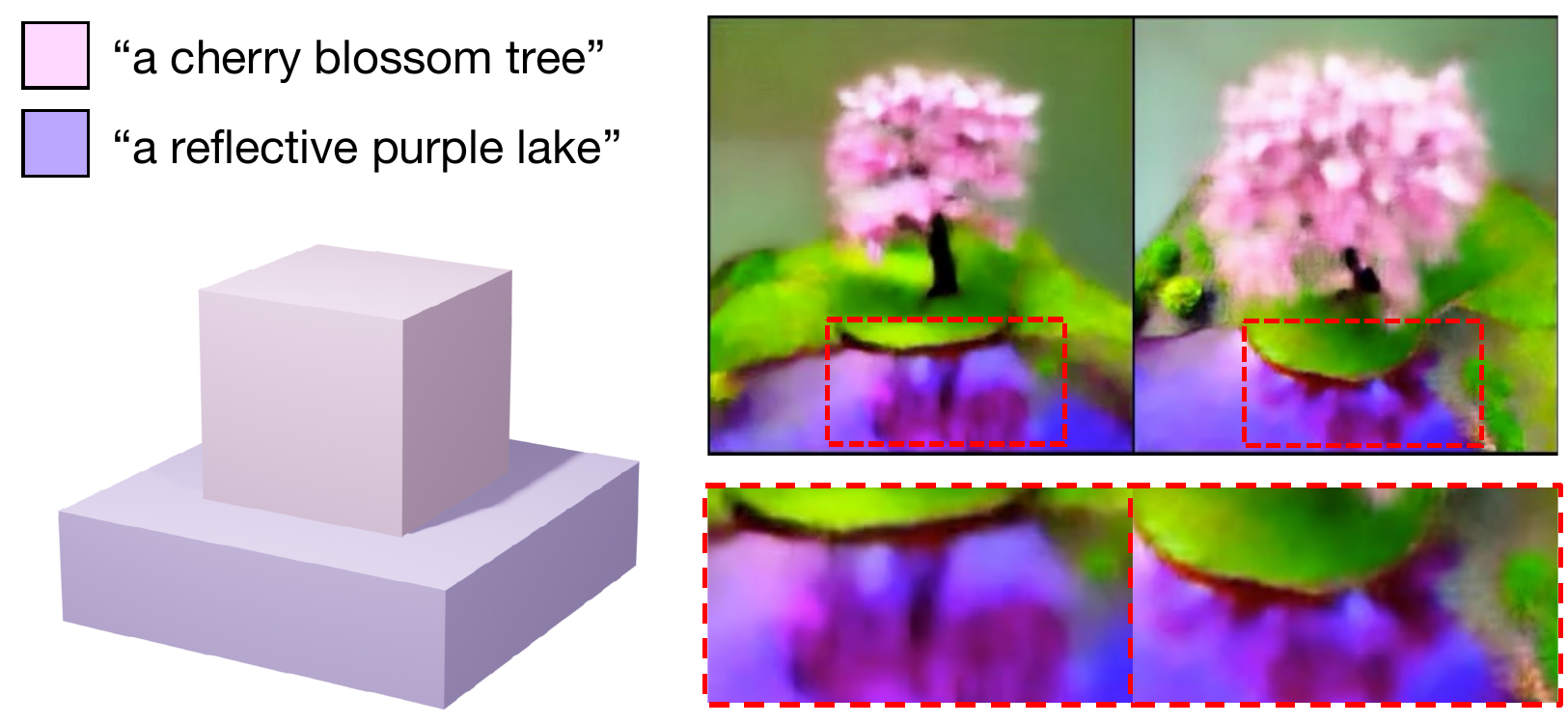}
    \caption{\textbf{Seamless transitions.} Our method is able to smoothly transition between scene components in different bounding boxes. In this example, we can see the reflection of the cherry blossom tree in the lake.}
    \label{fig:smooth}
    \vspace{-0.3cm}
\end{figure}
One option for compositional scene generation is to generate each scene component individually and then combine them manually afterwards. However, blending scene components together in a seamless manner takes considerable effort. Our method is able to blend individual objects with scene-level detail. As shown in Fig. \ref{fig:smooth}, although the cherry blossom tree and the reflective purple lake correspond to different bounding box regions, our method is able to generate reflections of the tree in the water. Such effects would not be present if each scene component were generated individually and then manually combined.

We also compare our method to a composable implementation of SJC using Composable-Diffusion \cite{Liu2022CompositionalVG}. However, this method fails to generate reasonable 3D scenes.

\paragraph{Quantitative results.} Following prior work \cite{Poole2022DreamFusionTU, Jain2021ZeroShotTO}, we evaluate the CLIP R-Precision, the accuracy of retrieving the correct input caption from a set of distractors using CLIP \cite{Radford2021LearningTV}, of our compositional method. Tab. \ref{tab:clip_eval} reports CLIP R-Precision values for rendered views of scenes shown in Fig. \ref{fig:baselines} using our compositional method and SJC with a single prompt. Our method outperforms the baseline across all evaluation methods.

\begin{table}[!ht]
\caption{CLIP R-Precision comparisons.}
  \centering
  \begin{tabular}{cccc}
  \toprule
    & \multicolumn3{c}{R-Precision $\uparrow$} \\
    \cmidrule{2-4}
    Method & B/32  & B/16  & L/14 \\
    \midrule
    Single Prompt (SJC) & 27..8 & 31.5  & 28.53\\
    Composed (Ours) & \textbf{38.6} & \textbf{54.3} & \textbf{29.8}\\
    \bottomrule
  \end{tabular}
  \label{tab:clip_eval}
    \vspace{-1em}
\end{table}

\begin{figure}[tb!]
    \centering
    \includegraphics[width=\linewidth]{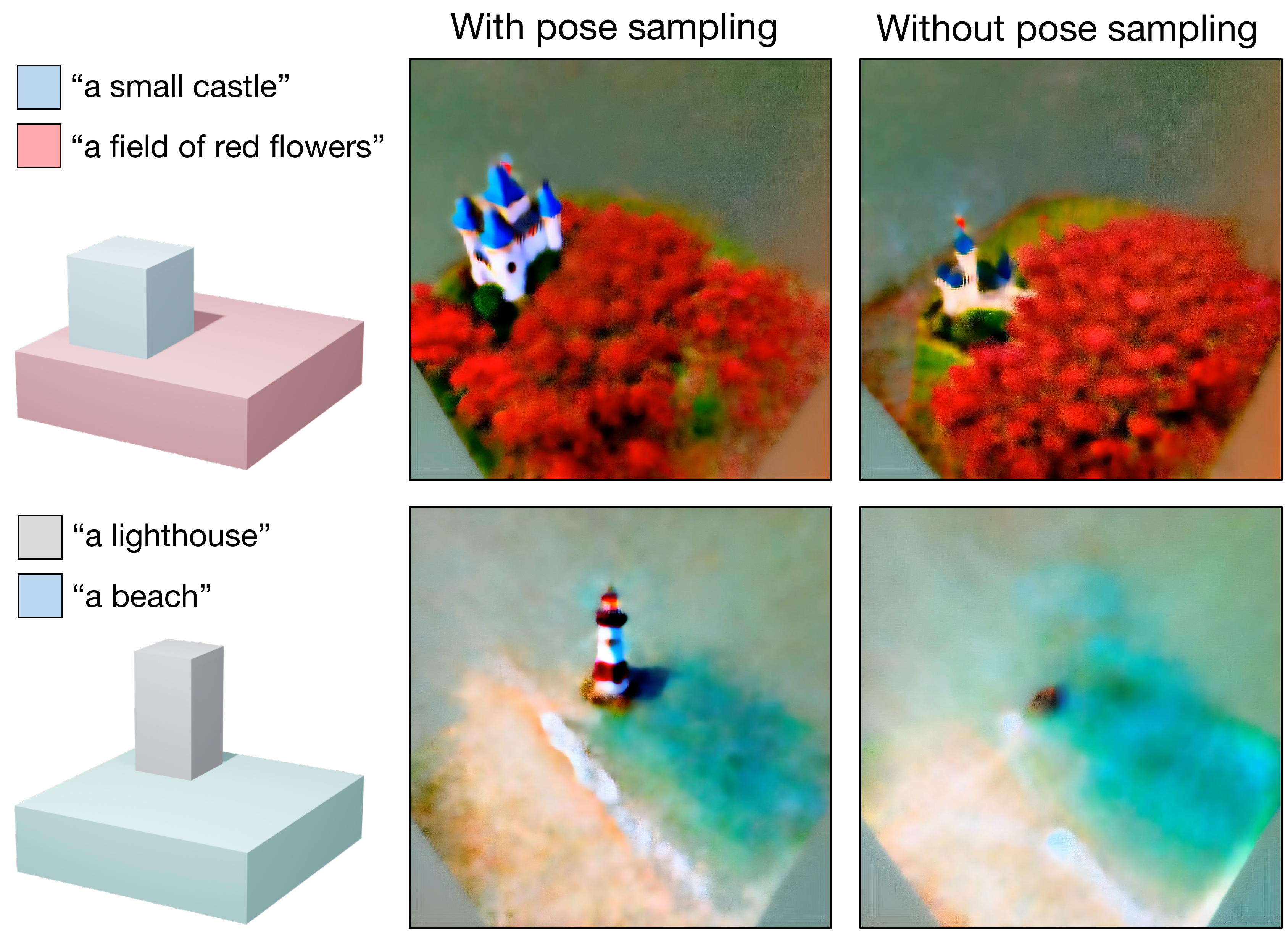}
    \vspace{-1em}
    \caption{\textbf{Ablation over object-centric sampling.} Without object-centric sampling, our method fails fully capture off-centered scene components.}
    \label{fig:ablations}
    \vspace{-1em}
\end{figure}

\paragraph{Ablations.} We found that object-centric camera pose sampling is essential for successful composition of multiple scene components. This is especially true for bounding boxes further away from the origin. We compare generations with and without object-centric pose sampling in Fig.~\ref{fig:ablations}. Note that our method tends to ignore certain scene components without object-centric sampling.

\paragraph{Speed evaluation.} Unless stated otherwise, all results were generated by running our method for 10000 denoising iterations with a learning rate of 0.05 on a single NVIDIA RTX A6000. Note that scenes with a higher number of distinct text prompts require a longer period of time to generate. Using SJC, generating scene components individually causes generation time to scale linearly with number of prompts. In contrast, our method can compose the same number of prompts in a shorter amount of time, as calls to the same diffusion prior conditioned on different text-prompts can be batched together. Table \ref{tab:speed_eval} shows generation times for SJC and our method for 3000 denoising iterations.

\begin{table}[ht!]
\caption{Generation times using SJC \cite{Wang2022ScoreJC} for individual prompts and composing multiple prompts using our method.}
  \centering
  \begin{tabular}{cccc}
  \toprule
    & & \# of prompts & \\
    \cmidrule{2-4}
    Method & 1  & 2  & 3  \\
    \midrule
    Individual (SJC) & 8 mins & 16 mins & 24 mins \\
    Composed (Ours) & 8 mins & 12 mins & 15 mins \\
    \bottomrule
  \end{tabular}

  \label{tab:speed_eval}
\end{table}

\section{Discussion and Conclusions}

Creating coherent 3D scenes is a challenging task that requires 3D design expertise and plenty of manual labor. Our method introduces a basic interface for creating 3D scenes without any knowledge of 3D design. Simply define bounding boxes for the desired scene components and fill in text prompts for what to generate in those regions.

\paragraph{Limitations and future work.} Although text-to-3D methods using SDS \cite{Poole2022DreamFusionTU, Wang2022ScoreJC, Lin2022Magic3DHT} have shown promising results, speed is still a limiting factor. While advances in image-diffusion-model sampling \cite{Lyu2022AcceleratingDM, Kong2021OnFS, Salimans2022ProgressiveDF, Watson2021LearningTE, Song2020DenoisingDI} have enabled the generation of high-quality results in dozens of denoising steps, SDS method still require thousands of iterations before a 3D scene can be learned. SDS-based methods are also limited by their reliance on unusually high guidance scales \cite{Poole2022DreamFusionTU}. A high guidance scale promotes mode-seeking, but leads to low diversity in the generated results. Concurrent works~\cite{BarTal2023MultiDiffusionFD, Jimnez2023MixtureOD} have shown other methods for controlling text-to-image diffusion synthesis with coarse segmentation masks. However, these methods require running a diffusion prior on multiple image patches before forming a single image, greatly increasing time needed to generate a single denoising step. In theory, these works could be applied in combination with our method, albeit with greatly increased time needed to generate a single 3D scene.

\paragraph{Ethical considerations.} Generative models, such as ours, can potentially be used for spreading disinformation. Such misuses pose a societal threat and the authors of this paper do not condone such behavior. Since our method leverages StableDiffusion \cite{Rombach2021HighResolutionIS} as an image prior, it may also inherit any biases and limitations found in the 2D diffusion model.

\paragraph{Conclusion.} Text-to-3D synthesis has recently seen promising advances. However, these methods mostly specialize in object-centric generations. Our method is an exciting step forward for 3D scene generation. Designing a 3D scene with multiple components no longer requires 3D modeling expertise. Instead, by defining a few bounding boxes and text prompts, our method can generate coherent 3D scenes that fit the input specifications.

\section*{Acknowledgements}
We thank Alex Bergman and Cindy Nguyen for valuable discussions and feedback on drafts. This project was in part supported by the Samsung GRO program. Ryan Po was supported by a Stanford Graduate Fellowship.

{\small
\bibliographystyle{ieee_fullname}
\bibliography{egbib}
}

\end{document}


\title{Compositional 3D Scene Generation using Locally Conditioned Diffusion:
Supplementary Materials}
\maketitle


\section{Implementation details}
\paragraph{CLIP R-precision.} We follow prior work \cite{Poole2022DreamFusionTU} in evaluating our compositional 3D generation method using CLIP R-precision. R-precision is the accuracy at which an image recovers the correct text caption among a set of distractors using CLIP \cite{Radford2021LearningTV} similarity. In our implementation we took 267 random text prompts from the DreamFusion gallery page and added 3 additional prompts for each scene: the composed prompt and the two individual prompts being composed. The results in the paper are an average taken over the 4 compositional scenes shown in Fig. 4.

\paragraph{3D implementation of Composable-Diffusion.} In our baseline, we compare our method to an implementation of Composable-Diffusion \cite{Liu2022CompositionalVG} applied to an SDS-based 3D generation pipeline (in this case SJC \cite{Wang2022ScoreJC}). Work in Composable-Diffusion has shown the potential of applying their global gradient composition method to a 3D diffusion pipeline such as Point-E \cite{Nichol2022PointEAS}. In our implementation, we simply replace the denoising step, with the composed denoising steps calculated using Composable-Diffusion. Results shown in Fig. 4 show that our implementation does not converge to a coherent scene. We provide additional results in Supplementary Fig. \ref{fig:composable} to show that our implementation works for other prompts. In our experiments, we found that our implementation of 3D Composable-Diffusion worked best when composing object-centric prompts, but failed to converge for prompts desceibing entire scenes. 

\begin{figure*}[h!]
    \centering
    \includegraphics[width=0.8\linewidth]{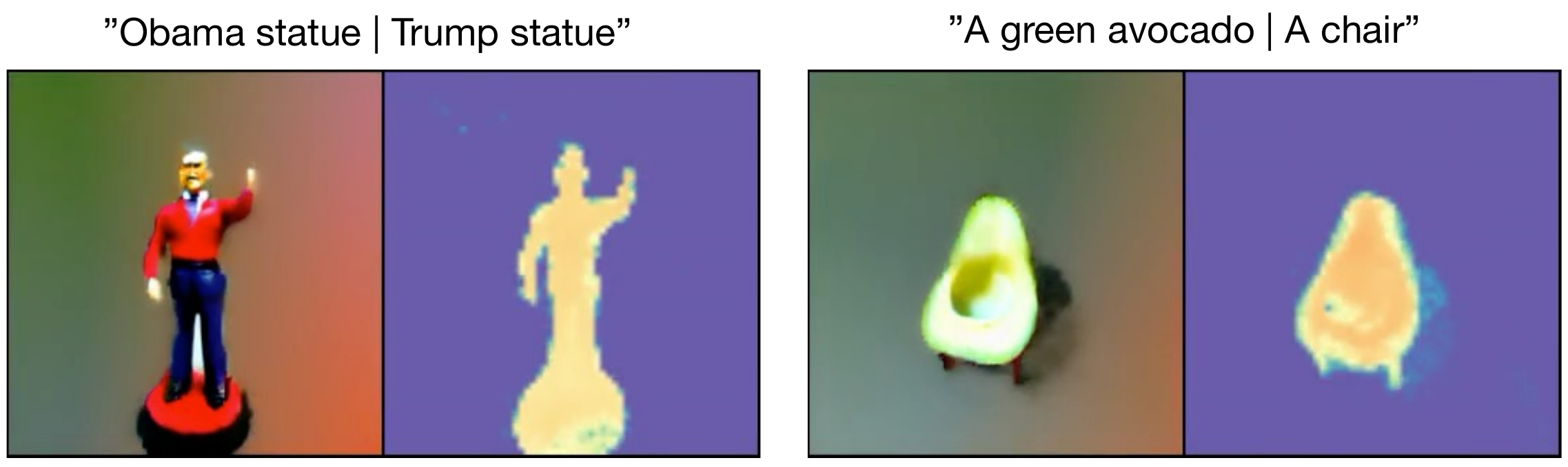}
    \caption{Additional generations from our implementation of Composable-Diffusion \cite{Liu2022CompositionalVG} applied onto SJC. Composable-Diffusion applied to an SDS-based pipeline tends to work better when composing object-centric prompts. Corresponding depth maps for each generation are shown on the right.}
    \label{fig:composable}
\end{figure*}

\paragraph{Object-centric camera pose sampling.} We found that object-centric camera pose sampling is essential for compositional 3D generation, especially for regions that are away from the center of the generation volume. SJC \cite{Wang2022ScoreJC} randomly samples camera angles facing the center (origin) of the VoxNeRF representation. During training, we sample the $i^{\text{th}}$ scene component with probability $p_i$. If a scene component is sampled, the camera pose is offset to face the center of the corresponding bounding box. We leave the values of $p_i$ as a hyper-parameter. We found that off-center scene components often require a higher sampling probability.

\paragraph{View-dependent prompting.} Following other SDS-based 3D generative models \cite{Poole2022DreamFusionTU, Wang2022ScoreJC, Lin2022Magic3DHT, Metzer2022LatentNeRFFS}, our method also relies on  view-dependent prompting. Depending on azimuth angle and elevation of the current camera pose, the prefixes 
''overhead view of", ''front view of", ''backside view of" and ''side view of" are added to the input prompts. As mentioned in prior works, this method alleviates the ''Janus problem", where multiple faces may appear on the same object. In addition to view-depending prompting, we found that adding the prefix ''a zoomed out photo of" and suffix ''highly detailed" too the input prompts would at times improve output generations.

\paragraph{Concept negation for locally conditioned diffusion. } Similar to Composable-Diffusion \cite{Liu2022CompositionalVG}, we show that our locally conditioned diffusion method also allows for concept negation. Recall the expression for noise estimation using classifier free guidance for some text prompt $y$,
\begin{equation}
    \hat{\boldsymbol{\epsilon}}_\phi(x_t, t, y) = \boldsymbol{\epsilon}_\phi(x_t, t, \emptyset) + s\Bigl(\boldsymbol{\epsilon}_\phi(x_t, t, y) - \boldsymbol{\epsilon}_\phi(x_t, t, \emptyset)\Bigr).
\end{equation}
Given a negation concept $y_n$, we can apply concept negation during noise estimation with the following,
\begin{equation}
    \hat{\boldsymbol{\epsilon}}_\phi(x_t, t, y, y_n) = \boldsymbol{\epsilon}_\phi(x_t, t, \emptyset) + s\Bigl(\boldsymbol{\epsilon}_\phi(x_t, t, y) - \boldsymbol{\epsilon}_\phi(x_t, t, y_n)\Bigr).
\end{equation}
We show qualitative comparisons of 2D generations with and without concept negation in Supplementary Fig. \ref{fig:negation}. Locally conditioned diffusion is applied to each semantic region as usual, but noise predictions are evaluated using concept negation (negated concept is the other text prompt in the semantic map). We found that using concept negation could help 2D locally conditioned diffusion generations to better adhere to input semantic maps. However, there was no notable difference when applied to the 3D setting. All 3D results shown in the main paper and supplementary materials do not use concept negation during generation.
\begin{figure*}[h!]
    \centering
    \includegraphics[width=\linewidth]{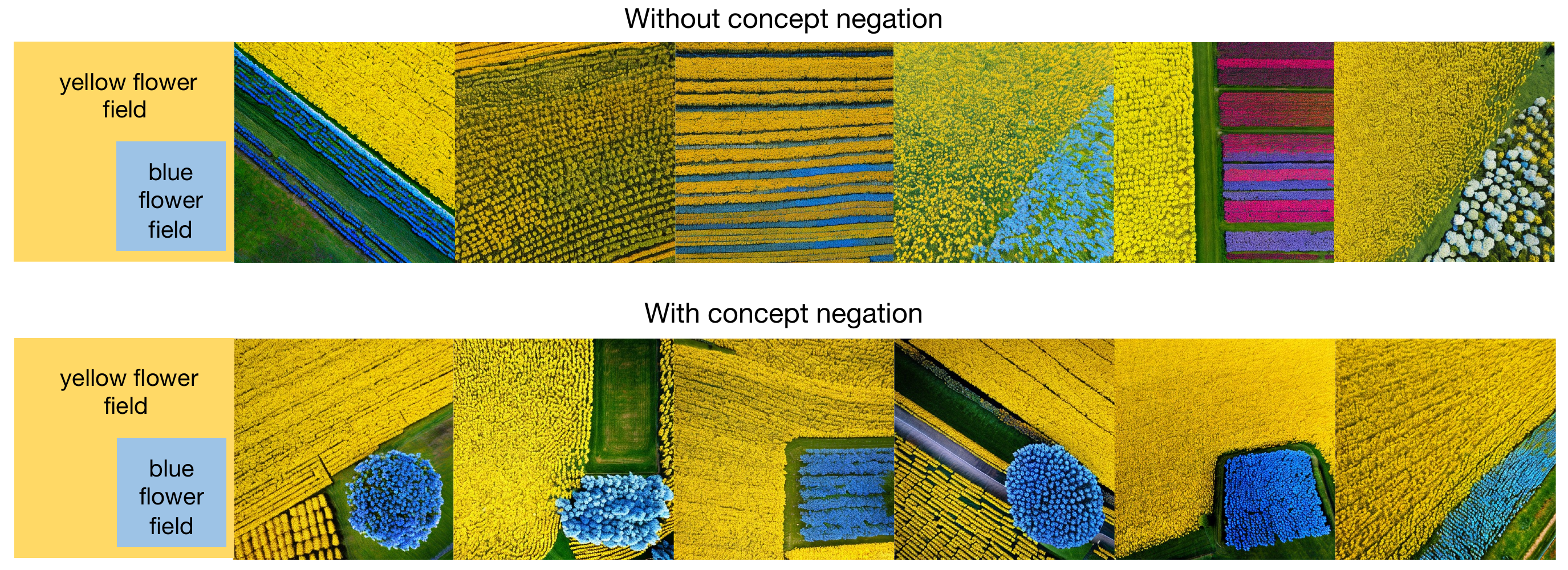}
    \caption{2D generations using locally guided diffusion with and without concept negation. With concept negation, our method adheres closer to the input segmentation masks.}
    \label{fig:negation}
\end{figure*}

\newpage

\section{Additional results}
\paragraph{Depth maps.} We show our results with their corresponding depth maps in Supplementary Fig. \ref{fig:depth}.

\begin{figure*}[h!]
    \centering
    \includegraphics[width=\linewidth]{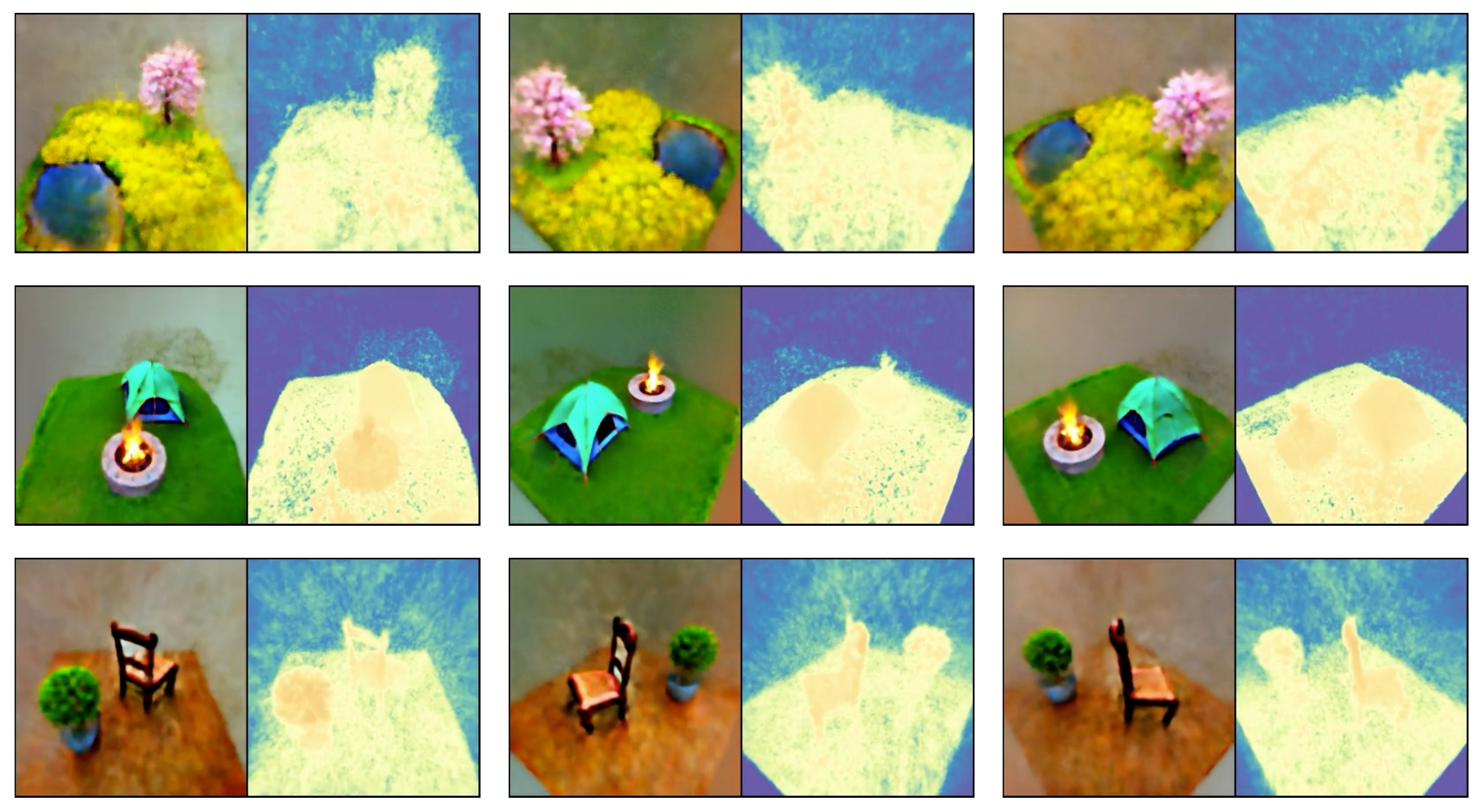}
    \caption{Main figure generations with corresponding depth maps.}
    \label{fig:depth}
\end{figure*}


{\small
\bibliographystyle{ieee_fullname}
\bibliography{suppbib}
}